\documentclass[letterpaper, 10 pt, conference]{ieeeconf}%
\usepackage[T1]{fontenc}%

\IEEEoverridecommandlockouts%

\let\labelindent\relax
\usepackage{enumitem}

\usepackage{amsmath,graphicx,algorithm,adjustbox,setspace}

\usepackage{xcolor}
\usepackage{enumitem}
\usepackage{hyperref}
\usepackage{multirow}
\usepackage{etoolbox,siunitx}
\robustify\bfseries
\usepackage{booktabs}
\usepackage{balance}
\usepackage{amssymb}
\usepackage{bm}
\usepackage{dsfont}
\usepackage{cite}
\usepackage{algorithm}
\usepackage[noend]{algpseudocode}
\usepackage{amsmath}
\usepackage[most]{tcolorbox}
\usepackage{mdframed}

\newenvironment{IEEEkeywords}{%
  \vspace{1em}%
\noindent\small\textbf{Keywords:} %
}{}

\let\OLDthebibliography\thebibliography
\renewcommand\thebibliography[1]{
  \OLDthebibliography{#1}
  \setlength{\parskip}{1pt}
  \setlength{\itemsep}{2pt plus 0.3ex}
}

\title{
KitchenVLM: Vision-Language Corrections for Robotic Execution of Human Tasks
}

\author{Kai Lu$^{1,2}$, Chenyang Ma$^{1,2}$, Ioannis Havoutis$^{1,*}$, Chiori Hori$^{2,*}$, Diego Romeres$^{2,*}$%
\thanks{This work is part of an internship at Mitsubishi Electric Research Laboratories (MERL), Cambridge, USA. }%
\thanks{$^{1}$: 
K. Lu, C. Ma, and I. Havoutis are with the University of Oxford, Oxford, UK. 
Email: {\tt\footnotesize kai.lu@oxfordrobotics.institute, ioannis@robots.ox.ac.uk, chenyang.ma@cs.ox.ac.uk}}%
\thanks{$^{2}$: C. Hori and D. Romeres are research scientists at MERL.
Email: {\tt\footnotesize chori@merl.com, diego.romeres@gmail.com}
}%
\thanks{$^{*}$: Equal advising. }%
        }%
        
\begin{document}

\bstctlcite{IEEEexample:BSTcontrol}

\maketitle

\begin{abstract}
In this paper, we present KitchenVLM, a multimodal framework for generating and optimizing executable robot actions from human instructional videos. While recent advances in video understanding and step generation have shown promising results, translating these steps into robot-executable actions remains challenging, particularly for complex, long-horizon tasks such as those in kitchen environments. These challenges arise from domain discrepancies between human videos and robotic settings, as well as mismatches between human actions and robot capabilities. KitchenVLM addresses these issues with an agentic feedback loop, where vision-language models (VLMs) serve as evaluators and planners to detect object or state mismatches, assess action feasibility, and generate corrective steps that refine execution. By incorporating keyframe selection, language-guided segmentation, and robot verification, KitchenVLM refines robotic plans to ensure contextual accuracy and executability. Through perspective evaluation and correction, our framework enhances the adaptability and robustness of robotic task execution in kitchen environments, advancing the integration of VLMs into robot learning and executable plan correction. Project page: \href{https://kl-research.github.io/kitchenvlm}{\textcolor{blue}{KitchenVLM}}.
\end{abstract}


\section{Introduction}
\label{sec:intro}

Learning robot action sequences from human videos is a key problem in embodied intelligence, with broad applications including domestic assistants and service robotics. Recent advances have shown that Vision-Language Models (VLMs) and Multimodal Large Language Models (MLLMs)~\cite{GPT-4o_openai, team2024gemini, ma2024spatialpin, hori_2025_icassp} demonstrate strong capabilities in interpreting instructional videos~\cite{LLaVA_Lin, Motonari_ICRA24_WS4Cooking} and generating structured textual steps, making them promising for the planning modules of multimodal robotic action generation frameworks.

\begin{figure}

  \centering
\vspace{2mm}
\includegraphics[width=\linewidth]{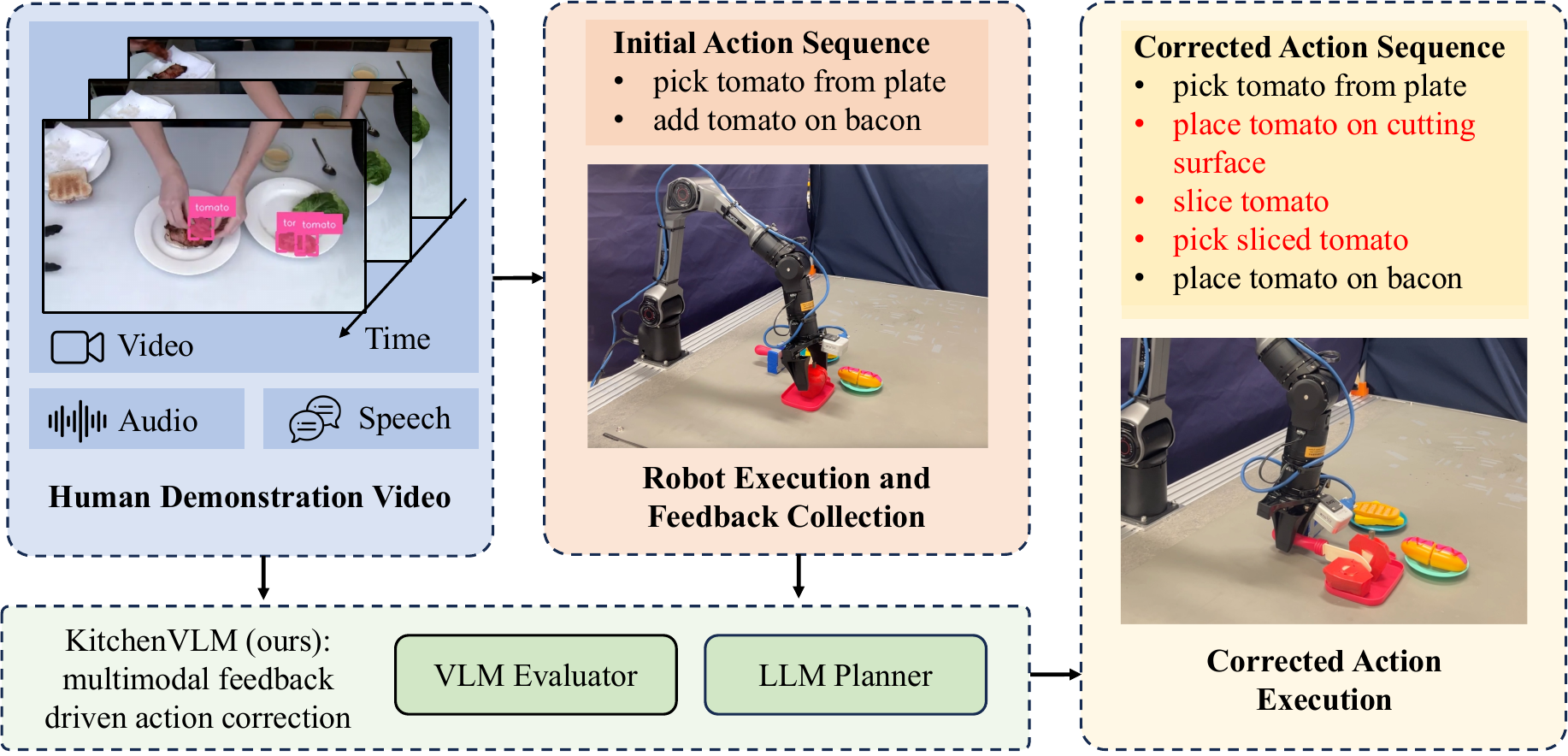}
 \vspace{-4mm}
  \caption{\textbf{Bridging Human-Robot Instruction Gaps via VLM Corrections.}
Given a human demonstration video, KitchenVLM extracts an initial action sequence, which may fail on the robot due to domain discrepancies (e.g., sliced vs. whole tomato). A VLM Evaluator detects mismatches between robot execution and the human video, and an LLM Planner corrects the sequence, generating a refined plan aligned with the robot’s capabilities and environment while preserving the original task intent.}
  \label{fig:teaser}
\vspace{-2mm}
\end{figure}

However, turning these steps into robot execution remains challenging. Monolithic approaches extend VLMs with control-level action decoders~\cite{team2024octo, brohan2023rt, KimPKXB0RFSVKBT24, pi05_vla_2025, ma2026furniturevla} or learn policies directly from demonstrations~\cite{MandlekarXWNWK021}, but they require large-scale robot data and computation. Hierarchical approaches instead translate VLM-generated steps into a predefined skill library or pretrained policies~\cite{IchterBCFHHHIIJ22, driess2023palmeembodiedmultimodallanguage, LiZX0SPZ24, Seedo_wang}, but they face three gaps: the environment in the video differs from the robot's workspace, human actions differ from robot capabilities, and the textual steps omit contextual cues of the demonstration (Fig.~\ref{fig:teaser}).

Most existing methods execute the extracted steps in a single pass through LLM-based translators~\cite{IchterBCFHHHIIJ22, liang2023code} or learned policies~\cite{team2024octo, MandlekarXWNWK021}, assuming that the robot's environment resembles the demonstration, and thus often fail in long-horizon tasks such as cooking, where existing systems rely on fixed task pipelines~\cite{yori_noh, MavrogiannisMA24} or skill-specific designs~\cite{Beltran-Hernandez24}. Failure-recovery methods edit plans with LLMs~\cite{RamanCIRMTP24}, reason about errors from multimodal observations~\cite{LiuBS23}, learn local recovery policies~\cite{vats2025recoverychaininglearninglocalrecovery, ma2026cyclevla}, or ask humans for guidance~\cite{shirasaka2023selfrecoverypromptingpromptablegeneral, ly2026inteliplan}, but they correct single steps against the robot's own environment and do not compare the execution with the human demonstration. Hence, it remains challenging to adapt a whole human-demonstrated sequence to a different robot and workspace.

To address these challenges, we propose KitchenVLM, a multimodal framework 
for domain-aware evaluation and correction of robot action sequences derived from human instructional videos. At the core of our framework is a feedback-driven replanning module, where VLMs and LLMs act as agents to compare the object states and task progressions in the human video with the robot’s observations after each action. When discrepancies are detected, e.g., a robot is instructed to pick a sliced tomato but only has access to a whole tomato, KitchenVLM generates and inserts corrective steps, such as slicing, to align execution with the intended goal. In addition to object-state discrepancies, our system handles semantic and logical issues such as incorrect object references, ambiguous instructions (e.g., interpreting \textit{add} as \textit{place} vs. \textit{rotate}+\textit{pour}), and logical inconsistencies (e.g., attempting \textit{place} before \textit{pick}).

KitchenVLM integrates three components: 1) action mapping, which grounds VLM-generated steps to robot-executable actions and objects; 2) robot execution with feedback collection, which logs observations and execution outcomes; and 3) VLM evaluation and LLM replanning, which compares robot observations with video keyframes and recomposes the corrections into a new action sequence.

In summary, our contributions are threefold: 
\begin{itemize} 
\item We propose a multimodal evaluation–correction pipeline that lets robots recover from domain gaps and incomplete demonstrations.

\item We develop a unified framework to map, detect, and correct mismatches in object state, action feasibility, and sequence logic.

\item We introduce a correction approach that refines action sequences and improves task success across diverse kitchen tasks, validated in simulated environments and on a real robot.

\end{itemize}

\section{Related Work}
\label{sec: related}

\subsection{Vision-Language Models and Robotic Action Generation}

\begin{figure}
  \centering \hspace{2mm} \includegraphics[width=0.85\linewidth]{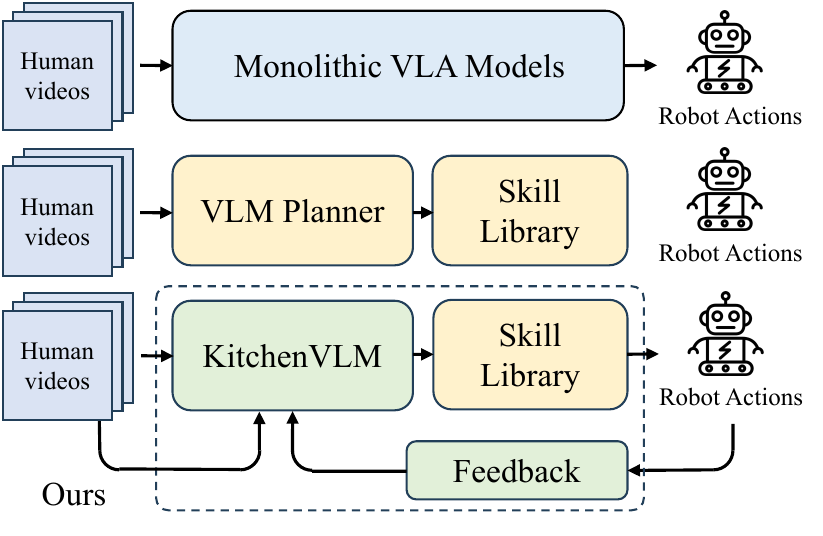}
  \caption{\textbf{Comparison of different methods for Vision-Language-based robotic action generation.} Unlike single-pass VLM planners or monolithic VLA models, KitchenVLM introduces a feedback loop to refine robot action sequences based on execution outcomes.
  }
  \label{fig:relatedwork}
\vspace{-4mm}
\end{figure}

Vision-Language Models (VLMs)~\cite{GPT-4o_openai, team2024gemini, SpatialVLM_Chen, ma2024spatialpin}
have demonstrated strong capabilities in connecting perception and language for guiding robotic actions.
OKAMI~\cite{LiZX0SPZ24} leverages VLMs for generating human-to-robot motion retargeting, while 
SayCan~\cite{IchterBCFHHHIIJ22} uses LLMs as high-level planners to compose robot behaviors 
from learned skills. These works exemplify VLM-based action reasoning rather than end-to-end execution, and our work follows this direction by using VLMs and LLMs as modular evaluators 
and planners for correcting high-level action sequences, adding a feedback loop from execution that single-pass planners and monolithic
models lack (Fig.~\ref{fig:relatedwork}).

Alternatively, end-to-end policy learning frameworks generate actions directly from demonstrations or 
multimodal inputs. RoboMimic~\cite{MandlekarXWNWK021} provides large-scale datasets and imitation policies 
for manipulation; Hydra~\cite{BelkhaleCS23} incorporates hierarchical imitation learning for long-horizon tasks; 
and OpenVLA~\cite{KimPKXB0RFSVKBT24} unifies vision, language, and action into a single model.  There are also works that improve generalization by decoupling skill learning from robot control or using reinforcement learning and fine-tuning~\cite{carta2024groundinglargelanguagemodels, lu2023decoupling, lu2024learning, deng2019deep}. 
Recently, GR00T\cite{bjorck2025gr00t} and $\pi_{0.5}$~\cite{pi05_vla_2025} present vision-language-action models with open-world generalization, enabling long-horizon and dexterous manipulation in new environments.

\subsection{Robotic Manipulation for Kitchen Tasks}

Cooking robots usually require more complex designs and operate under notable constraints due to kitchen environments. YORI~\cite{yori_noh} introduces a system with dual-arm robotic configurations for kitchen tasks, incorporating state machines to manage action sequences. SeeDo~\cite{Seedo_wang} follows a structured pipeline approach to video understanding and action mapping in kitchen environments, emphasizing task organization. 
Alternatively, FOON~\cite{paulius2016functional} focuses on knowledge graph-based representations of cooking tasks, while Cook2LTL~\cite{MavrogiannisMA24} explores logical task formulation using LTL. Research into modular robotic frameworks, such as MOSAIC~\cite{mosaic_wang}, provides insights into hierarchical control systems for cooking automation. Notably, SliceIt~\cite{Beltran-Hernandez24} and LAVA~\cite{BhaskarLSST24} address specific kitchen manipulation tasks like cutting and scooping, demonstrating the necessity for skill-specialized robotic interventions in kitchen environments.

\subsection{Failure Correction and Replanning in Robotics}
Failure correction in robotics is a crucial component in ensuring robust and reliable performance in real-world scenarios. Traditional approaches often involve re-execution of failed actions or predefined fallback strategies. However, recent research has emphasized the importance of integrating multimodal reasoning and feedback into the process~\cite{ma2026cyclevla}.
Recently, methods such as CAPE~\cite{RamanCIRMTP24} propose re-prompting strategies for replanning upon execution failures, aligning LLM-based approaches with classical task and motion planning (TAMP) paradigms. Similarly, 
REFLECT~\cite{LiuBS23} incorporates multimodal failure reasoning using visual–audio cues, while Vats et al.~\cite{vats2025recoverychaininglearninglocalrecovery} learn local recovery policies for multi-step manipulation.
These frameworks underscore the need for a multimodal and context-aware failure recovery mechanism that can adapt to various operational constraints.

\subsection{Video Understanding for Robotics}

Video understanding~\cite{LLaVA_Lin, Maaz2023VideoChatGPT} is a crucial component in enabling robots to learn from human demonstrations by extracting meaningful task sequences and structuring them into executable steps. Approaches such as VideoMME~\cite{videomme_fu} and AVBLIP \cite{Motonari_ICRA24_WS4Cooking, hori_2025_icassp} use video-to-text transformations to generate structured task representations for robotic execution. In robotic learning, the challenge lies in converting raw video demonstrations into a form that aligns with robotic capabilities and keeping the object states the same between the two domains. KitchenVLM builds upon these methodologies by incorporating a VLM Evaluator to assess discrepancies between human demonstrations and robotic execution, ensuring task feasibility and alignment within the robotic environment.

\begin{figure*}[t]
    \centering
\includegraphics[width=\textwidth]{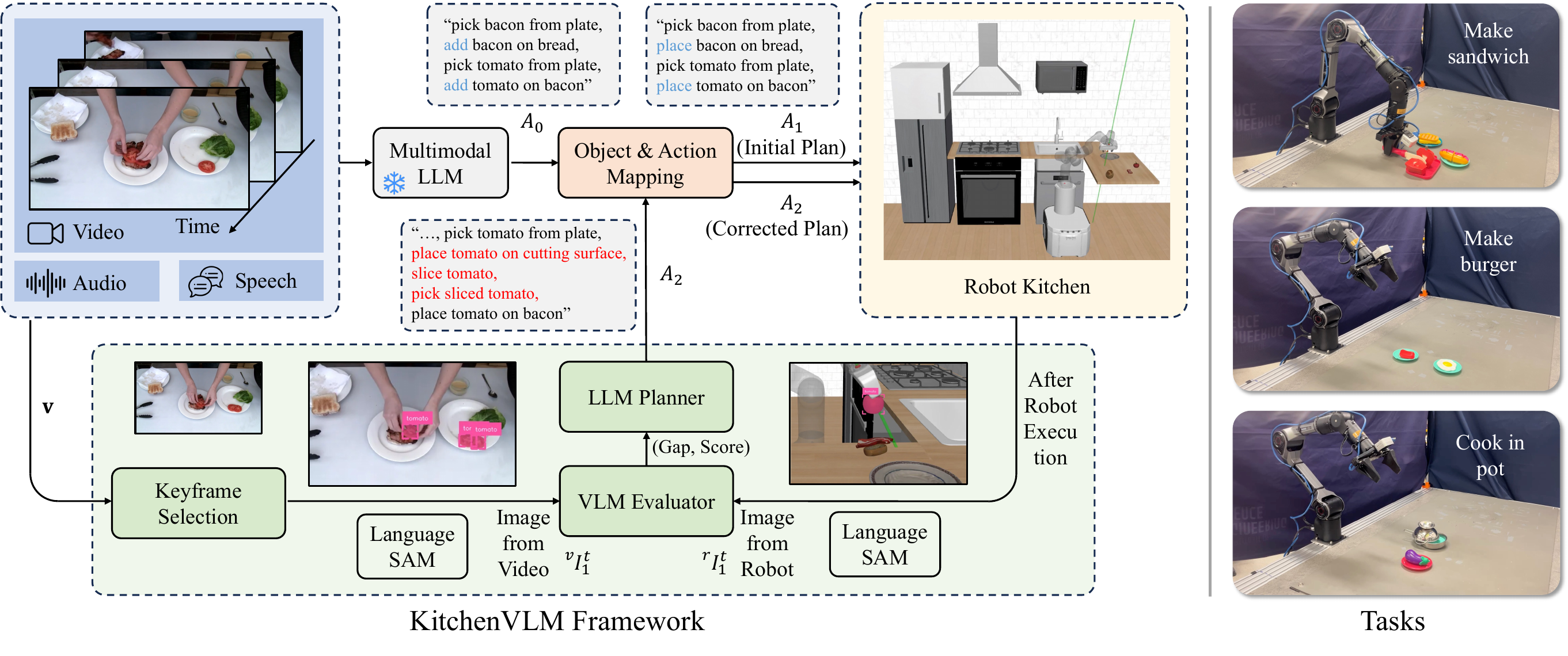}
\vspace{-8mm}
\caption{\textbf{Overview of the KitchenVLM framework and robot tasks.} 
Top: a multimodal LLM processes a human video to generate an initial action sequence, which is grounded into robot-executable actions through object and action mapping. 
Bottom: the correction stage, where keyframes from the video are compared with robot observations by a VLM evaluator to detect domain mismatches, and an LLM planner produces a corrected action sequence. 
KitchenVLM thus aligns human demonstrations with the robot's environment and skills.}
    \label{fig:frame}
\end{figure*}

\section{Problem Definition}

\label{sec:prob}
Human instructional videos represent a valuable resource for extracting high-level procedural instructions; e.g., the YouCook2~\cite{zhou2018towards} dataset contains demonstrations of cooking tasks in natural kitchen environments. 
Our goal is to generate a sequence of robot-executable actions that reproduces the task while handling embodiment and workspace differences between human and robot scenarios. Given a video, a pretrained multimodal LLM extracts a sequence of  steps
\begin{equation}
A_0 = \text{MLLM}(\mathbf{v}) = [a_0^0, a_0^1, \ldots, a_0^{T_0-1}],
\end{equation}
where each step is a tuple $a_0^t = (\omega_h, m_h, p_h, tr_h)$ of a human action verb $\omega_h \in \Omega_h$, a manipulated object $m_h \in O_h$, a prepositional relation $p_h$, and a target object $tr_h \in O_h$.
For example, the MLLM could be AVBLIP-2~\cite{hori_2025_icassp}, trained with human ground-truth labels, $A_0^*$, obtained from human annotators.
The robot offers a skill library $\Omega_r$ and an object set $O_r$, which generally differ from $\Omega_h$ and $O_h$.
Our goal is a robot-executable sequence $A_2 = [a_2^0, \ldots, a_2^{T_2-1}]$ that reproduces the task of $\mathbf{v}$ in the robot's workspace.

\section{KitchenVLM}
\label{sec:method}
KitchenVLM is our framework for bridging the domain gap between human demonstrations and robot execution (Fig.~\ref{fig:frame}). The task plan is corrected by comparing the feedback from the robot execution with the human video. The pipeline combines object–action mapping, execution feedback, VLM evaluation, and LLM replanning.

\vspace{-1mm}
\subsection{Object and Action Mapping}
\label{sec:obj_act_mapping}

Each human object is mapped to the most similar robot object,
$\text{map}_O(o_h) = \arg\max_{o_r \in O_r} \phi(o_h, o_r)$,
where $\phi(\cdot,\cdot)$ is a semantic similarity scored by an LLM, and each human verb is mapped to a robot skill by a deterministic lookup $\text{map}_\Omega: \Omega_h \to \Omega_r$ (e.g., \textit{add} $\mapsto$ \textit{place}). Applying both to every step gives the initial plan
\begin{equation}
\begin{aligned}
A_1 &= [a_1^0, \ldots, a_1^{T_0-1}], \quad\text{where} \\
a_1^t &= \big(\text{map}_\Omega(\omega_h), \text{map}_O(m_h), p_h, \text{map}_O(tr_h)\big).
\end{aligned}
\end{equation}
In the sandwich task (Fig.~\ref{fig:frame}), $A_1$ = [pick bacon from plate, place bacon on bread, pick tomato from plate, place tomato on bacon].

\subsection{Robot Execution with Feedback Collection}
\label{sec:exec}

$A_1$ is executed but differences in embodiment and environment, and the annotators' lack of robot-awareness, can make it fail through three sources of mismatch:
1)~\textit{object condition mismatch}: an object in the robot's workspace differs in type or state from the one in the video (e.g., a whole vs.\ a sliced tomato, a closed vs.\ an open pot);
2)~\textit{robotic action limitation}: a step is ambiguous or not in $\Omega_r$;
3)~\textit{logical inconsistency}: the sequence violates the robot's capabilities, e.g., placing an object before picking it with a single arm.
While executing $A_1$, the robot records an observation image ${}^rI^t$ and the execution outcome $e^t$ (e.g., what the gripper holds) after every step $a_1^t$.%

\subsection{Action Correction with VLM Evaluation}
\label{sec:correct}

After the robot execution, we can evaluate the discrepancies in results with the human demonstration. This forms the core of our framework’s ability to adapt the actions to robotic environments.

The VLM Evaluator compares the robot feedback  $^rI_1^t$ with the corresponding human video keyframe   $^vI_1^t = f_{\text{keyframe}}(\textbf{v})$. The keyframes are selected at equal temporal intervals. 
Both the instructional video and the robot view are cluttered, and we
direct the evaluator to the object of the current step: LangSAM \cite{langsam2023github} is queried with the
name of the manipulated object on both images, and the bounding box and label of the
highest-scoring detection are drawn on each image before it is passed to the VLM.

The VLM evaluator is then prompted to produce a domain-gap analysis and an action-step similarity score based on the images and current and next actions:
\begin{equation}
(g^t, s^t) = \text{VLM}_{\text{eval}}(a_1^t, a_1^{t+1}, {}^vI^t, {}^rI^t, e^t),
\end{equation}
where $g^t$ is a textual analysis of the domain gap (object coherence, motion feasibility, sequence logic) and $s^t \in [0,1]$ is a step similarity score. The window $(a_1^t, a_1^{t+1})$ lets the evaluator judge whether the current state permits the next step. The reason for comparing images rather than text alone is that the textual steps omit object states, e.g., whether the tomato is sliced, which only the video reveals.

The VLM output is a structured text that will be part of the prompt for the LLM planner, which operates in two stages: an action-level correction and a global sequence re-composition. 
In the first stage, the planner corrects each action window,
\begin{equation}
A_{\text{corr}}^t = \text{LLM}_{\text{replan}}(a_1^t, a_1^{t+1}, g^t, s^t),
\end{equation}
by inserting skills from $\Omega_r$, changing objects, or keeping $a_1^t$.

As we use action windows (current and next action in our case), the corrected sub-sequences, ${A}_{1,\text{corr}}^t$, are a partial plan and might have overlapping actions with another correction and they cannot be directly concatenated. 
Therefore, in the second stage all the corrected sub-sequences are collected into a set, and the LLM performs global sequence recomposition, reasoning over the set of corrected sub-sequences to resolve inconsistencies and redundancies. This leads to a coherent and complete action sequence:
\begin{equation}
A_2 = \text{LLM}_{\text{recompose}}\big([A_{\text{corr}}^0, \ldots, A_{\text{corr}}^{T_0-1}]\big).
\end{equation}

In the given example as shown in Fig. \ref{fig:frame}, the evaluator reports that the human places a tomato slice while the robot holds a whole tomato, and $A_2$ inserts actions [place tomato on cutting surface, slice tomato, pick sliced
tomato] before placing the slice.

\begin{figure}[t]
  \centering
  \includegraphics[height=2.75cm]{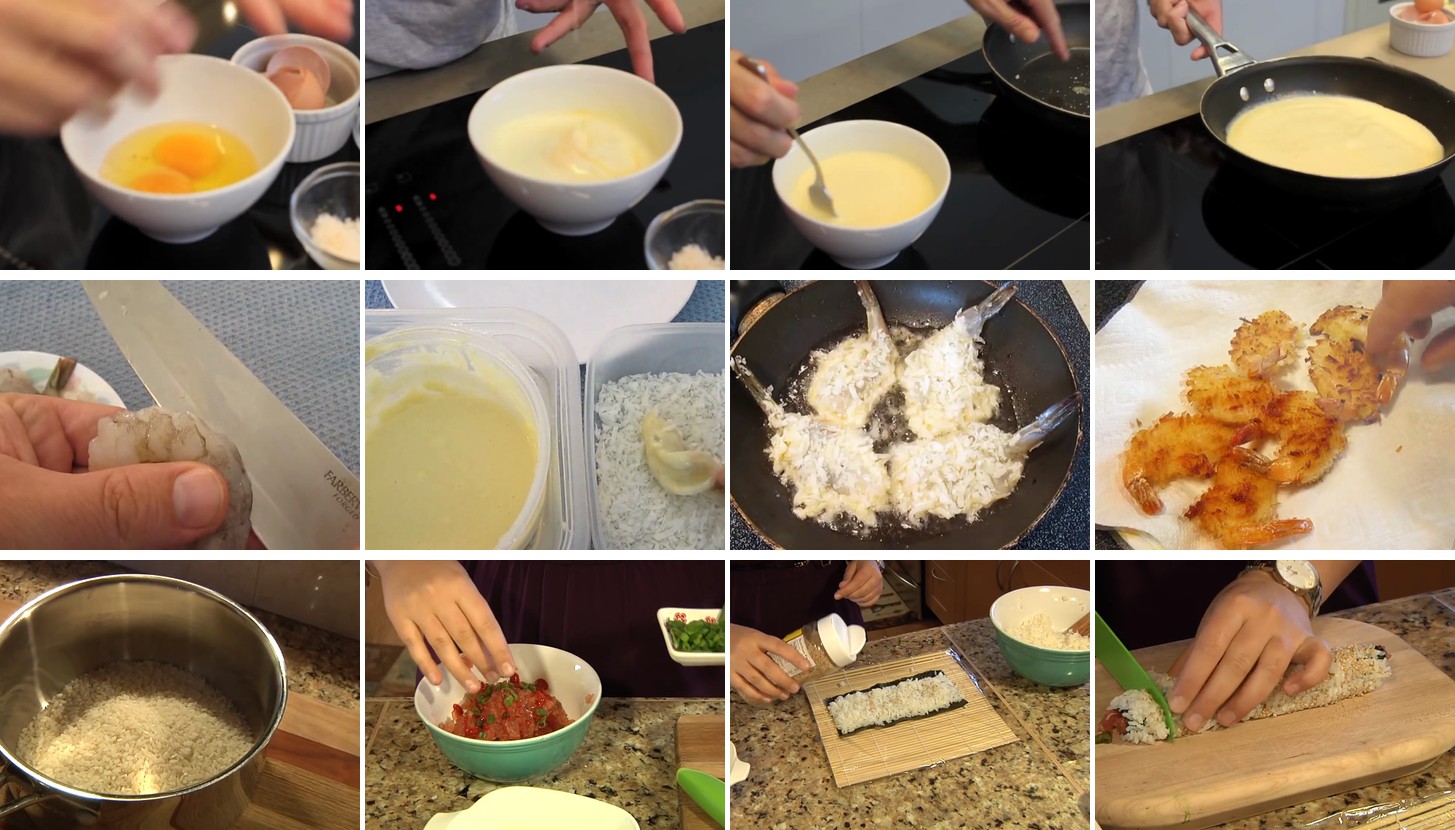}
  \hfill
  \includegraphics[height=2.75cm]{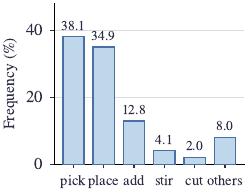}
  \caption{\textbf{Distribution of action verbs in the dataset.} Left: consecutive steps of three YouCook2 videos (one per row). Right: frequency of action verbs (\%), where \textit{others} covers the remaining 99 verbs.}
  \label{fig:action_freq}
\end{figure}

\begin{table}[t]
\centering
\caption{Judge scores of the initial plan $A_1$, the corrected plan $A_2$, and its ablations. $^*$: only successfully executable plans.}
\label{tab:ablation}
\setlength{\tabcolsep}{5pt}
\begin{tabular}{lcccc}
\toprule
Judge & $A_1$ & $A_2$ w/o S & $A_2$ w/o I & $A_2$ \\
\midrule
LLM $(G, A)$ & 0.45 & 0.25 & 0.56 & \textbf{0.67} \\
VLM $(G, I_0, A)$ & 0.35 & 0.30 & 0.58 & \textbf{0.65} \\
VLM$^*$ $(G, I_{0:n})$ & 0.36 & 0.46 & 0.53 & \textbf{0.64} \\
VLM$^*$ $(G, I_{0:n}, A)$ & 0.38 & 0.28 & 0.49 & \textbf{0.62} \\
\bottomrule
\end{tabular}
\end{table}

\section{Experimental Results}
\label{sec:exp}
\subsection{Experimental Setup}
\label{sec:sim_setup}

\textbf{Environment and Dataset.}
We use RoboCasa~\cite{nasiriany2024robocasa} with a mobile Franka Panda arm and a pseudo-suction gripper. Human steps come from YouCook2~\cite{zhou2018towards} with the Amazon Mechanical Turk (AMT) annotations of AVBLIP-2~\cite{hori_2025_icassp}. We use the 867 videos whose actions are covered by the robot's skills. YouCook2 verbs are long-tailed (Fig.~\ref{fig:action_freq}), illustrating the gap between human actions and robot skills.

\textbf{Implementation.}
GPT-4o~\cite{GPT-4o_openai} serves every LLM and VLM function in simulation, and LangSAM~\cite{langsam2023github} provides language-guided segmentation. The skill library is $\Omega_r$ = \{pick, place, move\_to, press, turn\_on, cut\} with 4 prepositional relations and 153 RoboCasa objects ($|O_h| = 1338$). The mapped human verbs are $\Omega_h$ = \{pick, place, add, cook, put, fry, take, dip, press\}. A wrist camera gives a hand-centric view comparable to the videos.

\textbf{Evaluation Metrics.}
1)~\textit{LLM judge}: an LLM scores a plan $A$ given the task goal $G$, $\text{Judge}_{\text{LLM}}(G, A)$, on the scale Reject $=0$, Borderline Reject $=0.25$, Borderline $=0.5$, Borderline Accept $=0.75$, Accept $=1$.
2)~\textit{VLM judge}: a VLM additionally sees the initial scene image $I_0$, or all step images $I_{0:n}$, with the same scale.
3)~\textit{Predicates}: for three dishes we define task-level predicates (e.g., ``tomato is sliced'') and report the fraction fulfilled.
4)~\textit{Human evaluation}: AMT workers correct the generated plans; we report BLEU-4, METEOR, and the task completion ratio (TCR), i.e., the fraction of clips whose plan needs no correction.

\textbf{Real-Robot Setup.}
We run the same pipeline on an AgileX PiPER 6-DoF arm with a parallel gripper. Two Intel RealSense D405 cameras provide an overhead and a wrist view, and the evaluator receives the overhead image after every step. The skill library is $\Omega_r$ = \{pick, place, cut\}; the skills are pre-defined and performed in shared ready poses, so that the planner can reorder skills. Before each replay, a symbolic gripper check (holding or empty) provides the execution outcome $e^t$. Toy food replaces real ingredients, and the object mapping substitutes, e.g., potato$\rightarrow$eggplant and cheese$\rightarrow$sausage. On the real robot, all LLM and VLM functions use GPT-6 Sol \cite{openai2026gpt6sol} with default sampling.

\begin{table}[t]
\centering
\caption{Predicate fulfillment of $A_1$ and $A_2$ on three dishes.}
\label{tab:predicate_eval}
\begin{tabular}{lcccc}
\toprule
Plan & Burger & Sandwich & Fried chicken & Avg. \\
\midrule
$A_1$ (initial) & 0.52 & 0.53 & 0.44 & 0.50 \\
$A_2$ (ours) & \textbf{0.61} & \textbf{0.63} & \textbf{0.57} & \textbf{0.61} \\
\bottomrule
\end{tabular}
\end{table}

\textbf{Real-Robot Tasks.}
We select three YouCook2 clips covering two sources of mismatch in Sec.~\ref{sec:exec}:
1)~\textit{Burger} (logical inconsistency): the annotated steps place the cheese before picking it;
2)~\textit{Pot} (object condition): the human puts potatoes into an open pot, whereas the robot's pot is closed by a lid;
3)~\textit{Sandwich} (object condition): the human stacks tomato slices, whereas the robot has a whole tomato, a knife, and a cutting board.

\textbf{Real-Robot Protocol.}
For each task, we execute $A_1$ without correction in 10 physical trials, and an operator marks task success and the task predicates. Corrections are then computed offline from the saved observations of these trials, so that all variants see identical robot data: \textit{full} (textual steps, robot images, and human keyframes), \textit{w/o I}, \textit{w/o S}, and \textit{state-only} (execution outcome only, in the spirit of CAPE~\cite{RamanCIRMTP24}). Full and w/o I are sampled three times per trial, the others once. As in Table~\ref{tab:predicate_eval}, we report the predicate fulfillment of each plan assuming perfect execution, where a step whose gripper precondition fails is skipped as on the robot, and the correction success, i.e., whether $A_2$ contains the required fix. On the real robot, the evaluator prompt additionally receives the list of scene objects $O_r$ (without it, the VLM mistakes the lidded toy pot for an open bowl), a checklist of the robot's state, and, in the full method, a textual summary of the human keyframe; LangSAM is not used.

\begin{table}[t]
    \centering
    \caption{Quality of Generated Action Sequence. \\TCR: Task Completion Ratio}
    \label{tab:result}
    \resizebox{\linewidth}{!}{
    \setlength{\tabcolsep}{6pt}
    \begin{tabular}{lcccccc}
        \toprule
                   & \multicolumn{3}{c}{\textbf{All}} & \multicolumn{3}{c}{\textbf{Aligned}}  \\
        \cmidrule(lr){2-4}\cmidrule(lr){5-7}
                       &  \textbf{BLEU-4} & \textbf{METEOR} & \textbf{TCR} & \textbf{BLEU-4} & \textbf{METEOR} & \textbf{TCR}\\
        \midrule
            Ours        & 0.705 & 0.503 & 0.28 & 0.720 & 0.526 & 0.29 \\
            Ours w/o S  & 0.230 & 0.278 & 0.03 & 0.211 & 0.274 & 0.03 \\
            Ours w/o I  & 0.633 & 0.485 & 0.23 & 0.643 & 0.500 & 0.25 \\
        \bottomrule    
    \end{tabular}
    }
\end{table}

\begin{table}[t]
\centering
\caption{Predicate fulfillment on the real robot, assuming perfect execution (mean over 10 trials; full and w/o I: 3 corrections per trial).}
\label{tab:real}
\begin{tabular}{lcccc}
\toprule
Plan & Burger & Pot & Sandwich & Avg. \\
\midrule
$A_1$ (initial)   & 0.00 & 0.00 & 0.25 & 0.08 \\
$A_2$ w/o S       & 0.50 & 0.00 & 0.50 & 0.33 \\
$A_2$ state-only  & \textbf{1.00} & 0.00 & 0.33 & 0.44 \\
$A_2$ w/o I       & \textbf{1.00} & 0.97 & 0.25 & 0.74 \\
$A_2$ (ours)      & \textbf{1.00} & \textbf{1.00} & \textbf{0.62} & \textbf{0.88} \\
\bottomrule
\end{tabular}
\end{table}

\subsection{Effectiveness of Feedback-Driven Correction}
Table~\ref{tab:ablation} shows that both judges rate the corrected plan $A_2$ higher than the initial plan $A_1$. The gain of the LLM judge (0.45$\rightarrow$0.67) indicates better task-level reasoning, while the gain of the VLM judge (0.35$\rightarrow$0.65), which also sees the scene, indicates better consistency with the robot's environment. The gain persists under the stricter VLM$^*$, which scores only executable plans.

\subsection{Predicate-Based Evaluation}
Predicates check the object states that the task requires. Assuming perfect execution of $A_2$ to isolate the planning quality, $A_2$ fulfills more task predicates on every dish (Table~\ref{tab:predicate_eval}), from 50\% to 61\% on average, recovering steps omitted by annotators.
This evaluation provides task-level insight beyond textual similarity and demonstrates KitchenVLM’s ability to produce semantically complete sequences even when human demonstrations are partial or ambiguous.

\begin{figure}[t]
    \centering
    \includegraphics[width=\linewidth]{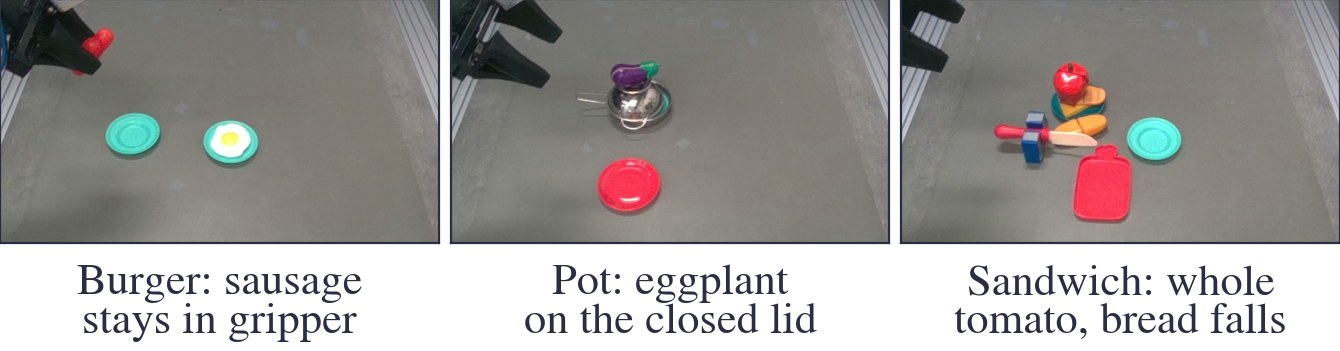}
    \vspace{-6mm}
    \caption{\textbf{Failures of the uncorrected plan $A_1$ on the real robot.}
Burger: the annotated steps place the sausage before picking it, so the \textit{place}
is refused (empty gripper) and the sausage picked afterwards is never placed. Pot: the eggplant is released onto
the closed lid. Sandwich: the whole tomato is stacked on the bacon, and the top
bread falls off.}
    \label{fig:real_fail}
\end{figure}

\begin{figure}[t]
    \centering
    \includegraphics[width=\linewidth]{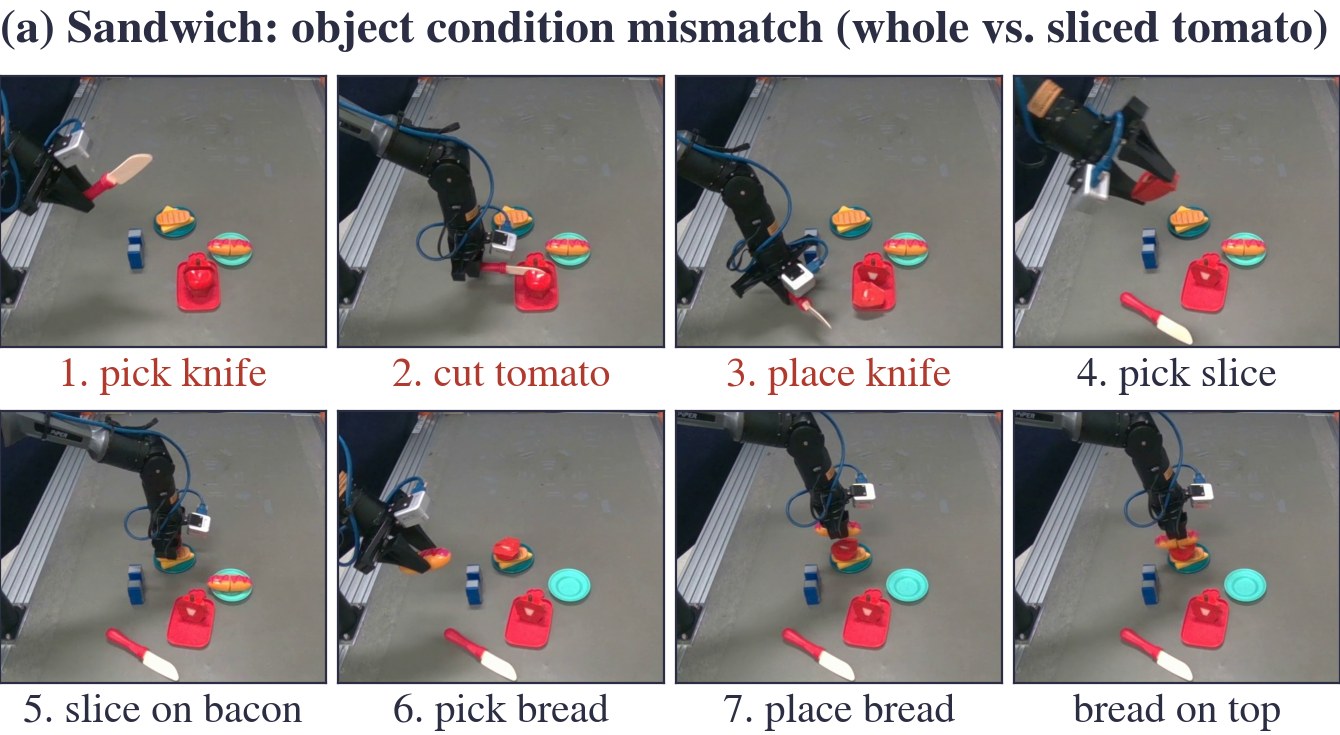}\\[-2.2mm]{\color{black!30}\rule{\linewidth}{0.4pt}}\\[0.4mm]
    \includegraphics[width=\linewidth]{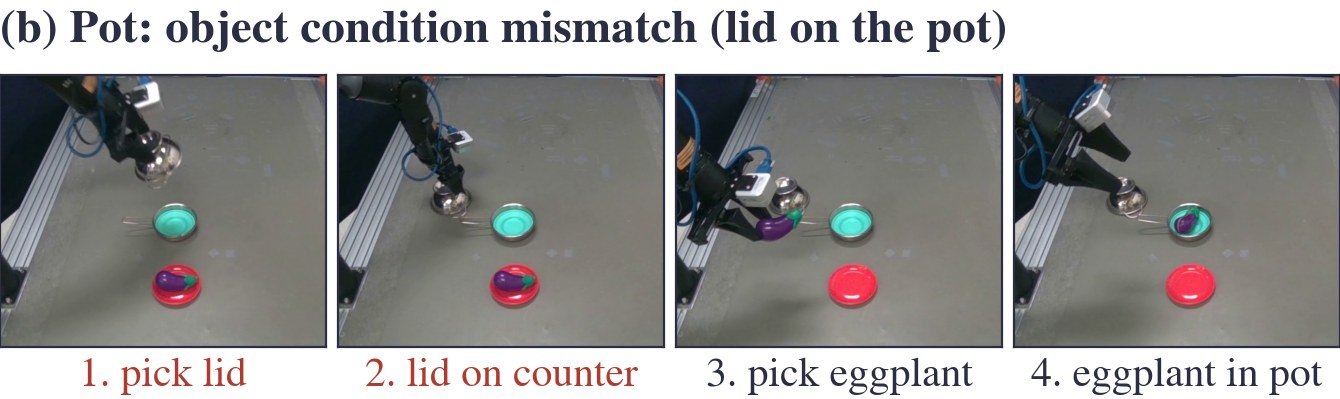}\\[-2.2mm]{\color{black!30}\rule{\linewidth}{0.4pt}}\\[0.4mm]
    \includegraphics[width=\linewidth]{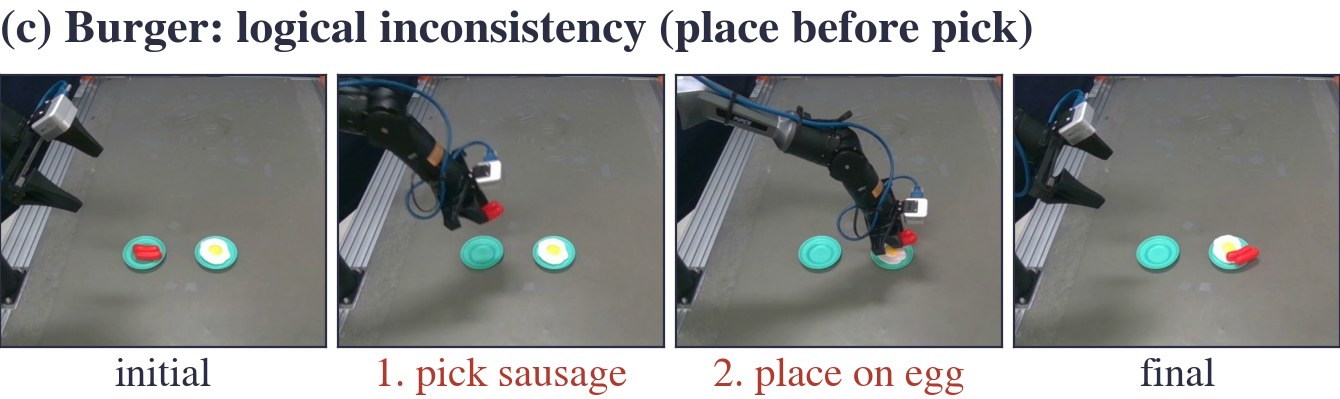}
    \caption{\textbf{Executing the corrected plans $A_2$ on the real robot.} Red: steps
inserted or reordered by KitchenVLM. (a) The tomato is cut before a slice is stacked,
so the top bread stays in place. (b) The lid is removed before the eggplant is placed
in the pot. (c) The sausage is picked before it is placed on the egg.}
    \label{fig:real_a2}
\end{figure}

\subsection{Human Evaluation and Importance of Keyframe Quality}

\begin{table*}[t]
    \centering
    \small
    \caption{Comparison of Initial and Corrected Action Sequences}
    \vspace{-0.5em}
    \label{tab:example}
    \resizebox{.99\linewidth}{!}{
    \begin{tabular}{lll}
        \toprule
       \textbf{ Video Id  }  & \textbf{Sequence Type}    & \textbf{Action Sequence} \\
        \midrule
        \multirow{2}{*}{{s99K\_WyajB8}}
			    & Initial Sequence        & pick onion, place onion to pan  \\
                \cmidrule(lr){3-3}
        	 & Corrected Sequence (Ours)  &  
             pick onion,
      \textcolor{red}{place onion to cutting board,
      pick knife, cut onion,
      place knife,
      pick onion,}
      place onion to pan,
      \textcolor{red}{press start}
      \\	
      \midrule
        \multirow{2}{*}{{TFJ6oR89Vb8}}
			    & Initial Sequence        &  pick egg from plate,
      add egg on pan \\
                \cmidrule(lr){3-3}
        	 & Corrected Sequence (Ours)  &  
             pick egg from plate,
      \textcolor{red}{crack egg into bowl,
      whisk egg mixture,
      pour egg mixture into pan}
      \\	
        \midrule
        \multirow{2}{*}{vZ7Pz9jM7zk}
			    & Initial Sequence        & pick chicken breast, place chicken breast in egg mixture  \\
                \cmidrule(lr){3-3}
        	 & Corrected Sequence (Ours)  &   pick chicken breast,
      place chicken breast in egg mixture,
      \textcolor{red}{press chicken breast in egg mixture}\\	
      \midrule
      \multirow{3}{*}{Xp2HNiLabRI}
			    & Initial Sequence        & pick butter stick from plate,
      place butter stick on bread,
      pick bread from plate,
      place bread on pan \\
      \cmidrule(lr){3-3}
        	 & Corrected Sequence (Ours)  & pick butter stick from plate,
      place butter stick on bread,
      \textcolor{red}{pick knife from drawer,
      spread butter on bread,} \\
      
      &&pick bread from plate,
      place bread on pan,
      \textcolor{red}{press start button on stove} \\	
        \midrule
      \multirow{3}{*}{8lqdPpg3w08}
			    & Initial Sequence        & pick egg from plate,
      place egg on grill,
      pick bottled\_water,
      add bottled\_water on grill \\
      \cmidrule(lr){3-3}
        	 & Corrected Sequence (Ours)  & 
              \textcolor{red}{pick spatula,
      pick egg with spatula,}
      place egg on grill,
      \textcolor{red}{place spatula back},
      pick bottled\_water,
      \textcolor{red}{move\_to grill,}
      \\ 
      &&
      \textcolor{red}{pour bottled\_water on grill,
      place bottled\_water on table}\\	
        \bottomrule
    \end{tabular}
    }
    \label{tab:action_compare}
\end{table*}

AMT workers corrected the plans generated by the full method, w/o S, and w/o I, and we use their corrections as ground truth. The full method scores highest in every metric (Table~\ref{tab:result}). Without textual steps, TCR drops to 0.03, as the text provides the order and structure of the task; without keyframes the drop is milder, as the robot images already ground most steps. All metrics improve on the 357 clips with aligned keyframes, showing that keyframe selection matters.

\subsection{Ablation Study}
Removing textual steps (w/o S) or human keyframes (w/o I) lowers every judge score (Table~\ref{tab:ablation}), so both inputs are needed. The drop is larger without textual steps (LLM: 0.67$\rightarrow$0.25) than without keyframes (0.67$\rightarrow$0.56), consistent with Table~\ref{tab:result}.

\subsection{Qualitative Comparison and Visualization}

The corrected plans insert missing prerequisites (e.g., \textit{place onion to cutting board, pick knife, cut onion}) and expand ambiguous steps into feasible ones (e.g., \textit{add egg} $\rightarrow$ \textit{crack egg into bowl, whisk egg mixture, pour egg mixture into pan}). 
The planner is instructed to prefer the skills in $\Omega_r$ but may propose a further
skill when the library is not sufficient. Such steps, e.g., \textit{crack}, \textit{whisk}, and \textit{spread}, cannot be executed by the current robot; they indicate which skill is missing, and the predicate evaluation counts them under its perfect-execution assumption.

Fig.~\ref{fig:real_a2} shows the execution of the corrected plans on the real robot, and Table~\ref{tab:action_compare} illustrates the different action sequences before and after the correction loop.

\subsection{Real-Robot Experiments}
\label{sec:real}

We further evaluate KitchenVLM on a physical robot to test whether the corrections transfer to real execution.

\textbf{Failures of the initial plan.}
Without correction, $A_1$ fails in all 30 physical trials; even under perfect execution, it fulfills only 0.00, 0.00, and 0.25 of the task predicates (Table~\ref{tab:real}). Fig.~\ref{fig:real_fail} shows the failures: in Burger, the first \textit{place} is refused as the gripper is empty, and the sausage stays in it; in Pot, the eggplant lands on the lid; in Sandwich, the whole tomato is stacked and the top bread falls off.

\textbf{Effectiveness of correction.}
KitchenVLM raises the predicate fulfillment from 0.08 to 0.88 on average and outperforms every ablation. Its corrections contain the required fix in 30/30, 30/30, and 15/30 cases for Burger, Pot, and Sandwich. Without textual steps (w/o S), no correction contains the fix; the partial scores of w/o S come from wrong plans that happen to satisfy single predicates.

\textbf{Feedback required by each task.}
The three tasks require increasingly rich feedback. Burger is solved by the execution outcome alone (state-only: 1.00), as the refused step directly reveals the wrong order. Pot requires the robot image, which shows the lid (state-only: 0.00, w/o I: 0.97). Sandwich is the only task where the missing information, that the tomato must be sliced, is visible only in the human video: without the keyframe, no correction inserts the cutting steps (0/30), whereas the full method does so in 15/30.

We visualize one corrected plan execution per task (Fig.~\ref{fig:real_a2}). All steps were executed, operator-marked predicates were satisfied, and the VLM judges in Table~\ref{tab:ablation} prefer the executed $A_2$ over $A_1$. The text-only LLM judge scores $A_1$ and $A_2$ equally on Pot and Sandwich, as the mismatch is only visible in the scene.

\section{Limitations and Future Work}
Our implementation is limited by the skills and objects available for the robot and therefore restricts KitchenVLM to a single corrective loop. Additional loops would require actions not present in the current skill library. The method itself, however, naturally supports multiple corrective loops.

For future work, we will explore KitchenVLM's generalization to other long-horizon manipulation domains beyond kitchen environments. Besides, we plan to investigate more robust keyframe selection and grounding mechanisms to further improve the reliability of feedback-based plan correction. We will also extend it to multiple correction loops and more skills.

\section{Conclusion}
\label{sec:concl}

In this work, we proposed KitchenVLM, a multimodal framework that bridges human instructional videos and robotic execution through feedback-driven plan correction. By comparing the robot's observations with keyframes of the human video, KitchenVLM detects mismatches in object state, action feasibility, and sequence logic, and inserts corrective steps into the plan. Experiments in simulation and on a real robot show that the corrected plans are more complete and executable than the initial ones, moving toward robots that learn from human videos and refine the action plans.

\bibliographystyle{IEEEtran}
\bibliography{mybib_arxiv}

\end{document}